\documentclass[pdflatex,sn-vancouver-num]{sn-jnl}
\usepackage{graphicx}%
\usepackage{multirow}%
\usepackage{amsmath,amssymb,amsfonts}%
\usepackage{amsthm}%
\usepackage{mathrsfs}%
\usepackage[title]{appendix}%
\usepackage{xcolor}%
\usepackage{textcomp}%
\usepackage{manyfoot}%
\usepackage{booktabs}%
\usepackage{algorithm}%
\usepackage{algorithmicx}%
\usepackage{algpseudocode}%
\usepackage{listings}%

\theoremstyle{thmstyleone}%

\theoremstyle{thmstyletwo}%

\theoremstyle{thmstylethree}%

\begin{document}
\title[CMHL for emotion Classification]{CMHL: Contrastive Multi-Head Learning for Emotionally Consistent Text Classification}

\author*[1]{Menna Elgabry}
\email{mennatullah.yasser@msa.edu.eg}

\author[1]{Ali Hamdi}

\author[2]{Khaled Shaban}

\affil[1]{Faculty of Computer Science, October University for Modern Sciences and Arts (MSA), Giza, Egypt}
\affil[2]{Computer Science and Engineering Department, College of Engineering, Qatar University, Doha, Qatar}

\abstract{
Textual Emotion Classification (TEC) is one of the most difficult NLP tasks. State of the art approaches rely on Large language models (LLMs) and multi-model ensembles. In this study, we challenge the assumption that larger scale or more complex models are necessary for improved performance. In order to improve logical consistency, We introduce CMHL, a novel single-model architecture that explicitly models the logical structure of emotions through three key innovations: (1) multi-task learning that jointly predicts primary emotions, valence, and intensity, (2) psychologically-grounded auxiliary supervision derived from Russell's circumplex model, and (3) a novel contrastive contradiction loss that enforces emotional consistency by penalizing mutually incompatible predictions (e.g., simultaneous high confidence in joy and anger). With just 125M parameters, our model outperforms 56x larger LLMs and sLM ensembles with a new state-of-the-art F1 score of 93.75\%  compared to (86.13\%-93.2\%) on the dair-ai Emotion dataset. We further show cross domain generalization on the Reddit Suicide Watch and Mental Health Collection dataset (SWMH), outperforming domain-specific models like MentalBERT and MentalRoBERTa with an F1 score of 72.50\% compared to (68.16\%-72.16\%) + a 73.30\% recall compared to (67.05\%-70.89\%) that translates to enhanced sensitivity for detecting mental health distress. Our work establishes that architectural intelligence (not parameter count) drives progress in TEC. By embedding psychological priors and explicit consistency constraints, a well-designed single model can outperform both massive LLMs and complex ensembles, offering a efficient, interpretable, and clinically-relevant paradigm for affective computing.}

\maketitle

\section{Introduction}

Human behavior, cognition, and social interaction are significantly influenced by emotions \cite{bib1}. They affect how decisions are made, how people communicate with each other, and how events are interpreted \cite{lerner2015emotion}. Large amounts of emotionally charged textual data are produced via social media posts, consumer reviews, discussions, emails, and online forums as digital communication becomes more and more integrated into daily life. Textual Emotion Classification (TEC) is a crucial task in natural language processing (NLP) that has applications in mental health analysis \cite{mohammad-etal-2018-semeval}, personalized recommendation systems \cite{GUNDERSEN2025114540}, customer experience monitoring, and human–computer interaction \cite{HCI}.

TEC is still one of the most difficult areas of emotion recognition, despite its significance. Textual input relies solely on linguistic signals, which are frequently unclear or implicitly transmitted, and lacks prosody, tone, body language, and visual clues when compared to modalities like speech or facial expressions \cite{wu-etal-2025-multimodal}. Figurative language, sarcasm, irony, metaphors, cultural allusions, and changing terminology all contribute to these difficulties by often masking a text's actual emotional content. Additionally, standard emotion datasets, such as the popular dair-ai Emotion dataset \cite{saravia2018emotion}, have limited size and substantial class imbalance, which makes it challenging for models to acquire strong representations for emotional categories that are underrepresented.

Lexicon-based techniques, traditional machine learning classifiers, and deep learning architectures based on pretrained language models have all been used in TEC systems throughout history. Because of their powerful language understanding capabilities, transformer-based models like BERT and RoBERTa have taken the lead \cite{Chutia2024}. To reach competitive outcomes a recurring work is the use of either sophisticated ensembles of small language models (sLMs) or large language models (LLMs), which frequently include billions of parameters \cite{Kumar2024}. Increased capacity is a benefit of these methods, but they come at the expense of memory usage, computational complexity, and poor real-world deployment efficiency.

Marginal performance advantages over individual models have been shown in recent studies using sLM ensembles, although these benefits frequently necessitate sophisticated aggregation techniques and hundreds of millions of combined parameters \cite{ZhangYang2022MTLSurvey}. Similar to Qwen \cite{bai2023qwen}, Mistral \cite{jiang2023mistral}, Falcon \cite{almazzouei2023falcon}, Phi \cite{gunasekar2023textbooks}, and OpenLLaMA \cite{geng2023openllama}, these LLMs have been optimized for TEC and show good performance, but they have a drawback, especially in environments with limited resources. This presents a crucial research question: Despite having a lot fewer parameters, can a well-designed single model perform better than both ensemble architectures and massive LLMs?

\label{sect:introduction}
\begin{figure}[h]
    \centering
    \includegraphics[width=0.99\textwidth]{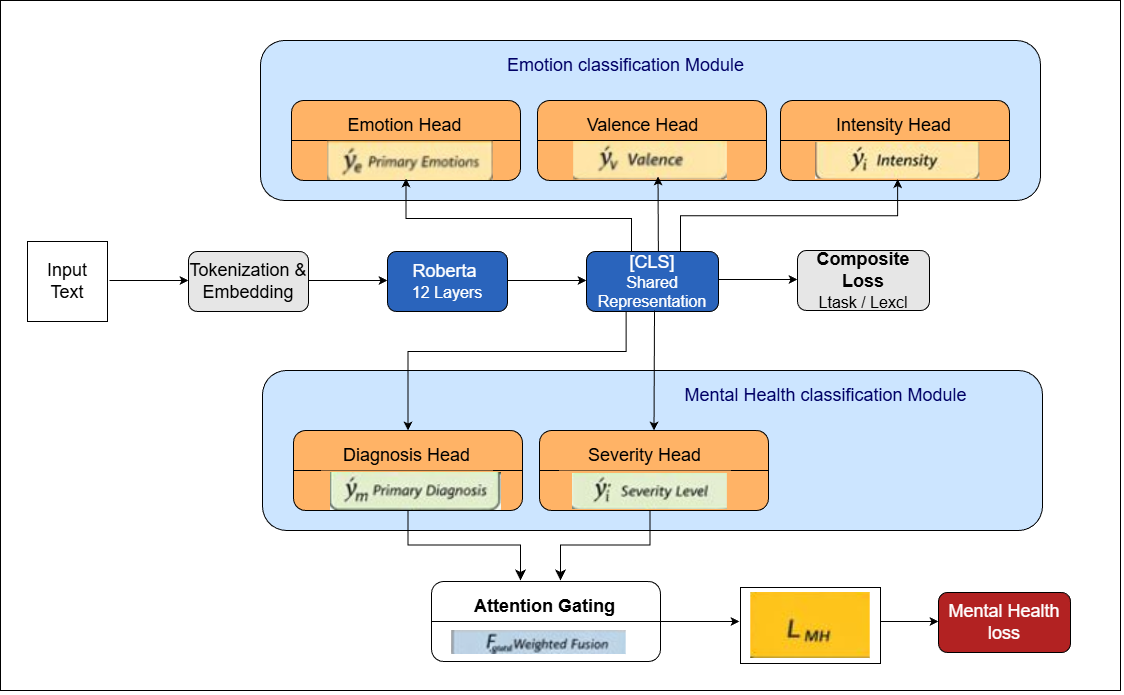}
    \caption{System Architecture}
    \label{fig:my_figure}
\end{figure}

In this paper, we dispute the widely held belief that larger or more complex systems are necessary for improved performance in TEC. We present a new single-model architecture as shown in figure1 \ref{fig:my_figure} based on RoBERTa that uses a unified multi-task learning framework to simultaneously predict valence polarity, emotion intensity, and core emotions. The architecture may learn common emotional representations across related tasks because of multi-head modeling, which enhances generalization and lessens misunderstanding between labels that are semantically similar.

Our work's main contribution is a contrastive contradiction loss, which is intended to penalize predictions that are logically inconsistent, like giving high confidence to both joy and anger at the same time. Such contradicting predictions are common in real-world systems, even as TEC models usually handle each output independently. Our contrastive term encourages the network to absorb logical links across emotional dimensions by directly modeling emotional incompatibility.

We conduct a thorough analysis using the dair-ai Emotion dataset and demonstrate that our model outperforms the original RoBERTa baseline, sLMs ensemble baselines (BERT–RoBERTa–DistilBERT–DeBERTa–ELECTRA hybrids), and several LLMs fine-tuned using LoRA \cite{Mao2024}, including Mistral and Falcon (7B-parameters models), by achieving a new state-of-the-art F1-score of 93.75\%. 

Our method only needs 125M parameters, setting a new standard for TEC efficiency and performance despite these significant performance gains. Our model's success emphasizes how crucial it is to explicitly model inter-label connections in fine-grained classification challenges. A straightforward but effective method for incorporating common psychological reasoning into a deep learning framework is the contrastive contradiction loss. This work demonstrates that better performance in TEC is more a result of design decisions that directly address the task's fundamental problems than it is of model scale.

\section{Related Work}
Emotion detection has received a lot of attention in recent years, and improvements have been made in many different areas, such as statistical models, transformer-based techniques, small large language models (sLMs), and multi-head output models. \cite{picard2003affective}.

\subsection{From Statistical Models to Contextual Embeddings with sLMs}
Early emotion identification systems depended on hand-engineered features and classical machine learning, which was essentially unimodal. This included n-gram models combined with classifiers like Support Vector Machines (SVMs) or Random Forests \cite{poria2017review}, bag-of-words representations, and lexicon-based sentiment scores (e.g., VADER). These approaches were interpretable, but their robustness in real-world situations was severely limited by their inability to comprehend word order, context, negation, sarcasm, and semantic nuance \cite{d2021challenges}.

Deep contextualized word embeddings and Small(er) Language Models (sLMs), particularly the Transformer-based BERT and its several variations (such as RoBERTa, ALBERT, and DistilBERT) \cite{Devlin2019BERT}, were the breakthrough. These models produce dynamic, context-sensitive representations for every token after being pre-trained on large corpora using goals like masked language modeling. sLMs showed a quantum leap in performance for textual sentiment and emotion classification when refined on emotion-labeled datasets \cite{barbieri2020tweeteval}. The capacity to identify long-range relationships and differentiate polysemous terms depending on context was their main breakthrough (e.g., separating "happy" in a joyful remark from its sarcastic usage).
The work of \cite{EmoBERTa}, who created EmoBERTa, a model designed especially for classifying emotions in clinical literature, is a clear example. In order to explicitly teach the model the language of emotional and psychological states, their methodology modified the RoBERTa architecture and pre-trained it using a sizable corpus of posts from mental health forums. After that, they refined it for emotion recognition using meticulously annotated clinical records. EmoBERTa significantly outperformed general-purpose BERT and other baselines thanks to this domain-adaptive pre-training, showing how sLMs can be tailored to comprehend complex, context-dependent emotions within a specific domain.
However, sLMs have intrinsic conceptual and architectural constraints. They are unable to simulate long-form discourse or extended conversational context, which is essential for monitoring emotional arcs \cite{to-etal-2024-deakinnlp}, because their context window is usually fixed and very short (e.g., 512 tokens for standard BERT). They function as unimodal text encoders at a deeper level. Despite their strength in language analysis, they are inherently unable to interpret or integrate the non-linguistic (such as gestures and facial expressions) and paralinguistic (such as tone and prosody) signals that make up human emotional communication \cite{zadeh2020multimodal}. The next evolutionary stage (LLMs) was driven by this unimodal bottleneck as well as their restricted world knowledge and reasoning capacity in comparison to larger models.

\subsection{ The Rise of Large Language Models (LLMs) and Their Dominance}
The GPT series, LLaMA, and PaLM\cite{Chowdhery2022PaLM} are examples of large language models (LLMs) that scale the Transformer architecture to previously unheard-of sizes (billions to trillions of parameters) and training data \cite{Touvron2023LLaMA, brown2020language}. This scale offers a number of important benefits that directly address the drawbacks of sLM:

\begin{itemize}
\item Extended Context \& Reasoning: By processing contexts of hundreds of thousands of tokens, models such as GPT-4 and Claude \cite{Bai2022ConstitutionalAI} enable sophisticated reasoning about emotional states throughout lengthy dialogues or narratives \cite{openai2023gpt4}.

\item In-Context Learning \& Instruction Following: LLMs can conduct emotion classification via few-shot prompting or instruction-tuning, adjusting to new emotion taxonomies or languages without parameter updates, in contrast to sLMs that need task-specific fine-tuning \cite{lu-etal-2024-emergent}.

    \item Inferring emotions from implicit or ambiguous utterances requires LLMs to have a broad awareness of social circumstances, cultural norms, and pragmatic clues. This is made possible by their extensive training corpus. \cite{shapira-etal-2024-clever}.
\end{itemize}

The work of \cite{sabour-etal-2024-emobench}, who carried out a thorough assessment of large language models on EmotionBench, a benchmark of 21 different emotion detection tasks, is an excellent example that methodically confirms these LLM advantages for complicated emotion analysis. In three settings—zero-shot, few-shot, and instruction-following—their technique compared models such as GPT-3.5 and GPT-4 against cutting-edge task-specific models (such as refined RoBERTa). The study showed that on tasks involving pragmatic interpretation and commonsense reasoning, like deriving emotions from abstract scenarios or unclear social media posts, instruction-tuned LLMs not only matched but frequently outperformed specialized models. Importantly, the performance discrepancy was greatest on the most challenging tasks, supporting the idea that a better, more comprehensive knowledge of nuanced affect results from the size and scope of LLM training data. This method demonstrates how a basic shortcoming of previous sLMs in tracking emotional dynamics during prolonged interactions is addressed by the long-context modeling and nuanced comprehension of LLMs.
As a result, on challenging affective computing tasks, LLMs have established new cutting-edge benchmarks. For instance, on complex emotion recognition datasets such as GoEmotions \cite{demszky2020goemotions}, instruction-tuned LLMs perform better than fine-tuned BERT models.

\subsection{ LLMs' Computational Cost Bottleneck}
The high computational cost of LLMs substantially restricts their practical deployment despite their powerful capabilities. Three crucial aspects are where this bottleneck appears:

First, memory footprint: Excluding memory for activations during inference, a model such as LLaMA 2-70B needs more than 140GB of GPU RAM simply to load in half-precision (FP16). This surpasses the capabilities of all but the most expensive and specialized hardware \cite{Touvron2023LLaMA}.

Second, inference latency: Due to the high latency caused by the autoregressive generation and large number of parameters, real-time or interactive applications (such as responsive tutoring systems and live conversational agents) are not practical \cite{tang2024computational}.

Third, energy consumption: The objectives of sustainable and scalable AI are at odds with the significant environmental impact and operational costs of training and inference for LLMs \cite{strubell2019energy}.

These limitations make monolithic LLMs unsuitable for high-throughput applications (social media analysis platforms), edge computing (mobile devices, IoT), and resource-constrained research environments. This has sparked a crucial shift in research: can a coordinated set of smaller, more effective models achieve equivalent performance if a single massive model is too expensive?

\subsection{A Strategic Synthesis of Effective Ensembles of Specialized sLMs}
This completes the story by returning to the employment of sLMs, but with an advanced new tactic. The objective is to build a collection of effective, specialized models whose combined intelligence is close to or greater than that of a single LLM while keeping the overall computational budget well below the LLM's specifications. Condorcet's Jury Theorem (CJT), which shows that the majority decision of numerous independent, reasonably competent voters can be more accurate than that of a single expert as long as voter errors are uncorrelated, serves as the theoretical foundation for this strategy \cite{lam1998economic}.

This translates to the following guidelines for creating efficient ensembles in machine learning:

First, variety through specialization: in accordance with Condorcet's Jury Theorem, numerous specialized sLMs are used to decrease error correlation. For example, research has shown that combining models with different architectural foundations—like BERT, RoBERTa, and ELECTRA—each refined on complementary emotional tasks (like comprehending sentiment, sarcasm, or cultural expressions) produces more robust emotion recognition than any one expert.\cite{yao-etal-2021-knowledge}. This method guarantees that a greater range of linguistic and pragmatic events are covered by the collective judgment of the ensemble.

Second, Architectural and Strategic Diversity: Recent research uses several model variations to minimize correlated error, which is crucial for CJT, in order to preserve efficiency. Distilled models (like DistilBERT, TinyBERT), parameter-efficient models (like ALBERT), and quantized versions of bigger bases are commonly included in ensembles \cite{Sanh2019DistilBERT, sun-etal-2020-mobilebert}. The main takeaway from this field of study is that variety needs to be strategic because the independence requirement of CJT cannot be met by merely replicating the same lightweight design. In order to effectively reduce correlated failures on difficult inputs like sarcasm or cultural nuance, the field has shifted toward purposefully heterogeneous ensembles where members differ in pre-training aims and architectural biases \cite{condorcet2023}.

Third, Dynamic \& Learnable Fusion: Averaging or simple voting are frequently not the best options. Research focuses on mixture-of-experts layers or meta-learners (a tiny neural network) that dynamically weight each specialized model's contributions according to the input sample \cite{yao-etal-2021-knowledge}. 

This groundbreaking research \cite{condorcet2023}, titled "Textual emotion detection with complimentary BERT transformers in a Condorcet’s Jury theorem assembly," specifically applies CJT ideas. An ensemble of BERT variants—two BERTs, two RoBERTas, and DistilBERT—are trained as the jury members using their technique. The authors suggest three CJT-based voting algorithms to guarantee independence, a crucial CJT requirement: Jury Classic (JC) for majority voting, Jury Adaptive (JA) with a juror memory system, and Jury Dynamic (JD), which employs reinforcement learning for dynamic juror selection in unsupervised circumstances.Their findings validate the use of CJT for reliable and effective emotion identification by confirming that all ensemble variations perform better than individual models with limited data.Although this work offers a useful application of CJT principles, there may be a methodological flaw with the independence requirement. For the "wisdom of the crowd" to be completely realized, CJT theoretically demands that juror errors be uncorrelated. However, because BERT and RoBERTa have comparable inductive biases, the authors' ensemble has two instances of each model, which runs the danger of producing correlated error patterns. More recent studies on ensemble design have placed a strong emphasis on this factor.\cite{yao-etal-2021-knowledge}.

\subsection{Multi-Head Output Models in Classification of Emotions}
The multi-head output model is an attractive architectural solution that strikes a compromise between extreme efficiency and specialization. Based on Multi-Task Learning (MTL), this paradigm uses several task-specific output heads with a single shared backbone encoder \cite{Liu2019MTDNN}. Instead of using separate, full-sized models for every emotional aspect (as in an ensemble), this architecture uses a strong, unifying feature extractor and assigns small, specialized parameters—the heads—to different jobs. Because of this, it is far more computationally and parameter-efficient than even an ensemble of sLMs.

Jointly learning linked emotional characteristics through a shared representation produces better and more resilient models, as research has repeatedly demonstrated. For example, a deep multi-task learning framework for combined multi-modal sentiment and emotion analysis was proposed by \cite{akhtar-etal-2019-multi}. They analyze text, audio, and visual inputs from video utterances using a single shared encoder, which is a collection of bi-directional Gated Recurrent Unit (GRU) networks. A contextual inter-modal attention method that calculates pairwise attention between modalities (such as text-visual and text-acoustic) is the main novelty. The most contributing characteristics are dynamically identified and weighted by this approach across contextual utterances in a conversation as well as across the various modes. Two separate, lightweight task-specific output heads then share the resulting unified representation: one with a sigmoid layer for multi-label emotion classification (anger, disgust, fear, happy, sad, surprise) and another with a softmax layer for binary sentiment classification (positive/negative). The model takes use of the interdependence of the tasks by training these heads together. For instance, understanding a "negative" sentiment improves the prediction of related emotions like "anger" or "sadness." This multi-head, multi-task architecture achieved state-of-the-art results for both sentiment and emotion analysis when tested on the CMU-MOSEI dataset, outperforming single-task baselines.

Thus, the multi-head architecture is the logical conclusion of the efficiency-specialization story: it provides a way to maintain the computational efficiency of a single model while achieving the specialization advantages of an ensemble (with each head concentrating on a different task like sentiment, sarcasm, or cultural nuance). The Condorcet-inspired demand for a variety of "experts" is neatly addressed by instantiating them as lightweight, divergent paths atop a shared knowledge base rather than as distinct models. Our work is directly motivated by this principle. We concentrate on modifying a single, strong, and effective encoder—RoBERTa—and enhancing it with a complex, multi-head output architecture rather than building an ensemble of several sLMs. With the lightweight footprint and simplified deployment benefits of a unified model, our design enables the model to specialize internally for complementing aspects of textual emotion, resulting in robust, nuanced classification.

\section{Methodology}
\label{sec:methodology}

Our two-tier approach is presented in this section: (1) a novel multi-head emotion classification framework with derived affective qualities, and (2) a validation research on mental health classification that shows architectural generalization. Auxiliary supervision\cite{sogaard2016deep}, adaptive feature integration, and multi-task learning\cite{ZhangYang2022MTLSurvey} are common ideas used by both components.

\subsection{Framework for Classifying Emotions}
By adding inferred affective qualities, this framework presents a novel multi-head emotion classification method that goes beyond categorical prediction. We automatically extract valence and intensity labels from pre-existing emotion annotations using Russell's circumplex model, allowing for richer affective representation learning without the need for more manual annotation. The design ensures psychological coherence between emotion predictions by combining multi-task learning with a unique composite loss function.
\label{subsec:emotion-framework}
\label{sec:methodology}

\subsubsection{Problem Formulation}
Our training dataset is represented by $\mathcal{D} = \{(x_i, y_i^e)\}_{i=1}^N$, where $x_i$ represents a textual input and $y_i^e \in \{1, \dots, C\}$ its primary emotion label from $C$ categories. Our goal is to develop a model $f_\theta: \mathcal{X} \rightarrow \mathcal{Y}^e \times \mathcal{Y}^v \times \mathcal{Y}^i$ that concurrently predicts intensity $y^i$ (high/low), valence $y^v$ (positive/negative/neutral), and emotion $y^e$. Importantly, the initial dataset only contains $y^e$ labels; Russell's circumplex model is used to construct $y^v$ and $y^i$. This allows multihead architectures to be designed that acquire richer, more structured affective representations from typical emotion-only annotations.

\subsubsection{Affective Attribute Derivation via Russell's Model}
\label{subsec:russell-derivation}
We automatically extract valence and intensity labels in accordance with Russell's affect theory \cite{russell1980circumplex}, which places emotions in a two-dimensional space of valence and arousal:

\begin{equation}
\begin{aligned}
& \text{Valence mapping: } \mathcal{V}: y^e \mapsto y^v \\
& y^v = 
\begin{cases}
\text{positive} & \text{if } y^e \in \{\text{joy, love}\} \\
\text{negative} & \text{if } y^e \in \{\text{sadness, anger, fear}\} \\
\text{neutral} & \text{if } y^e = \text{surprise}
\end{cases}
\end{aligned}
\end{equation}

\begin{equation}
\begin{aligned}
& \text{Intensity mapping: } \mathcal{I}: y^e \mapsto y^i \\
& y^i = 
\begin{cases}
\text{high} & \text{if } y^e \in \{\text{anger, fear, surprise, joy}\} \\
\text{low} & \text{if } y^e \in \{\text{sadness, love}\}
\end{cases}
\end{aligned}
\end{equation}

By establishing psychologically grounded supervision for auxiliary tasks, these mappings allow the model to learn more complex affective representations without the need for manual annotation.

\subsubsection{Multi-Head RoBERTa Architecture} \label{subsec:roberta-architecture} \label{sec:methodology}

We expand on the RoBERTa-base model \cite{liu2019roberta}, which has $H=12$ attention heads and $L=12$ transformer layers with hidden dimension $d=768$. Contextualized representations are obtained given an input sequence $x$.

\begin{equation}
\mathbf{H}^{(0)} = \text{Embedding}(x) \in \mathbb{R}^{n \times d}
\end{equation}

\begin{equation}
\mathbf{H}^{(l)} = \text{TransformerLayer}^{(l)}(\mathbf{H}^{(l-1)}) \quad \text{for } l=1,\dots,L
\end{equation}

where $n$ is the sequence length. The [CLS] token representation $\mathbf{h}_{\text{[CLS]}} = \mathbf{H}^{(L)}[0] \in \mathbb{R}^d$ serves as the aggregate sequence representation  following the standard transformer architecture \cite{vaswani2017attention}.

\paragraph{Task-Specific Heads} From $\mathbf{h}_{\text{[CLS]}}$, we compute:
\begin{align}
\hat{y}^e &= \text{softmax}(\mathbf{W}^e \mathbf{h}_{\text{[CLS]}} + \mathbf{b}^e) \in \mathbb{R}^C \\
\hat{y}^v &= \text{softmax}(\mathbf{W}^v \mathbf{h}_{\text{[CLS]}} + \mathbf{b}^v) \in \mathbb{R}^3 \\
\hat{y}^i &= \text{softmax}(\mathbf{W}^i \mathbf{h}_{\text{[CLS]}} + \mathbf{b}^i) \in \mathbb{R}^2
\end{align}
where $\mathbf{W}^e \in \mathbb{R}^{C \times d}$, $\mathbf{W}^v \in \mathbb{R}^{3 \times d}$, $\mathbf{W}^i \in \mathbb{R}^{2 \times d}$ are learnable projection matrices.

\subsubsection{Novel Composite Loss Function}
\label{subsec:composite-loss}

We suggest a multi-task learning goal that simultaneously optimizes all three prediction tasks while imposing a psychologically grounded constraint: a text should not be simultaneously categorized with high confidence for opposing emotional valences (e.g., high-probability "joy" and "sadness"). In order to achieve this, our composite loss incorporates a novel Adaptive Emotional Exclusivity Contrastive Loss together with a task-balancing term.
The primary balancing loss is defined as:
\begin{equation}
\mathcal{L}_{\text{task}} = \mathcal{L}_{\text{CE}}(\hat{y}^e, y^e) + \alpha_1 \mathcal{L}_{\text{CE}}(\hat{y}^v, y^v) + \alpha_2 \mathcal{L}_{\text{CE}}(\hat{y}^i, y^i)
\end{equation}

where $\alpha_1, \alpha_2 \in \mathbb{R}^+$ are hyperparameters governing auxiliary task contributions, and $\mathcal{L}_{\text{CE}}$ represents cross-entropy loss. Through thorough validation, we were able to determine the ideal values of $\alpha_1 = \mathbf{0.3}$ and $\alpha_2 = \mathbf{0.2}$ through extensive validation.

We include a contrastive loss term to penalize contradicting high-confidence predictions. For the $C$ fundamental emotion classes, let $P^e \in \mathbb{R}^C$ be the softmax probability vector. Sets of positive-valence emotion indices (such as joy and love) are defined as $\mathcal{P}$, and sets of negative-valence indices (such as sadness, wrath, and fear) as $\mathcal{N}$. The loss of exclusivity is:

\begin{equation}
\mathcal{L}_{\text{excl}} = \sum_{i \in \mathcal{P}} \sum_{j \in \mathcal{N}} \max(0, P^e_i + P^e_j - \tau_{ij})
\end{equation}

where $\tau_{ij} $ is a adaptive threshold that penalizes the sum of probability for opposing emotions above $\tau_{ij}$ but permits a modest margin of co-activation. The model is discouraged from giving mutually exclusive emotional states high probabilities by this approach.

A distance-based approach determines the adaptive threshold $\tau_{ij}$:

\begin{equation}
\tau_{ij} = \tau_0 + \alpha \cdot d(e_i, e_j)
\end{equation}
where $d(e_i, e_j)$ is a normalized affective distance between emotions $e_i$ and $e_j$ in a valence-arousal space, $\tau_0$ is a base threshold, and $\alpha$ is a scaling factor. This applies a psychologically informed penalty: closer emotions (like joy–excitement) receive a softer $\tau_{ij}$, allowing for a larger margin of co-activation, whereas emotions that are farther away in affective space (like joy–sadness) have a stronger $\tau_{ij}$, enforcing a tighter penalty.

Next, the overall composite loss is:
\begin{equation}
\mathcal{L}_{\text{total}} = \mathcal{L}_{\text{task}} + \lambda \cdot \mathcal{L}_{\text{excl}}
\end{equation}
where $\lambda$ controls the strength of the exclusivity constraint. We select $\lambda$ through validation experiments as $\lambda$ = 0.4.

\paragraph{Gradient Analysis} The gradient flow through shared parameters $\theta_{\text{shared}}$ becomes:
\begin{equation}
\nabla_{\theta_{\text{shared}}} \mathcal{L}_{\text{total}} = \nabla_{\theta_{\text{shared}}} \mathcal{L}_{\text{task}} + \lambda \cdot \nabla_{\theta_{\text{shared}}} \mathcal{L}_{\text{excl}}
\end{equation}
With the penalty scaled adaptively by the affective distance between emotion pairs, the $\mathcal{L}_{\text{excl}}$ gradient explicitly modifies the shared representations to decrease the simultaneous activation of antagonistic emotion logits. Compared to single-task emotion classification, this guarantees that the encoder learns representations that are both internally consistent with regard to emotional valence and discriminative for all three affective dimensions, resulting in a more complex and psychologically plausible understanding.
\subsubsection{Training Details}
With learning rate $\eta = \mathbf{2\times 10^{-5}}$, weight decay $\lambda = \mathbf{0.01}$, and linear warmup over $\mathbf{10}\%$ of training steps, we train with the AdamW optimizer. When needed, the batch size is set to $\mathbf{16}$ with gradient accumulation over $\mathbf{2}$ steps\cite{mikolov2012empirical}. There is a maximum sequence length of $\mathbf{256}$ tokens. The dair-ai emotion dataset with $\mathbf{20,000}$ samples from $C=\mathbf{6}$ emotion classes is used for training for $\mathbf{5}$ epochs.

\subsection{Validation: Mental Health Classification Framework}
We adapt the multi-head architecture for mental health classification in order to verify the generalizability of our method. This illustrates architectural adaptability while using severity-aware modeling to meet clinical demands. The framework uses an attention-based gating mechanism that dynamically weights feature contributions based on input characteristics to forecast both clinical severity levels and diagnostic categories.

\label{subsec:mental-health-validation}
\label{subsec:emotion-framework}
\label{sec:methodology}

\subsubsection{Architectural Generalization}
We adapt our multi-head strategy to mental health categorization on the Reddit Suicide Watch and Mental Health Collection dataset ($\mathbf{SWMH}$) to verify its broad applicability. Here, we forecast intensity levels $y^{i'} \in \{0, 1, 2\}$ that correlate to low, medium, and high clinical severity as well as primary mental health disorders $y^m \in="1, \dots, M\}$.

\subsubsection{Extended Multi-Head Architecture}
\label{subsec:extended-architecture}
The architecture extends our emotion model with task-specific modifications:

\begin{align}
\hat{y}^m &= \text{softmax}(\mathbf{W}^m \mathbf{h}_{\text{[CLS]}} + \mathbf{b}^m) \in \mathbb{R}^6 \\
\hat{y}^{i'} &= \text{softmax}(\mathbf{W}^{i'} \mathbf{h}_{\text{[CLS]}} + \mathbf{b}^{i'}) \in \mathbb{R}^5
\end{align}

where:
\begin{itemize}
    \item $\mathbf{W}^m \in \mathbb{R}^{M \times 768}$ and $\mathbf{b}^m \in \mathbb{R}^M$ are learnable parameters for the primary mental health classification head, producing probability distributions over $5$ diagnostic categories. This head directly corresponds to the main mental health classification task.
    
    \item $\mathbf{W}^{i'} \in \mathbb{R}^{3 \times 768}$ and $\mathbf{b}^{i'} \in \mathbb{R}^3$ parameterize the intensity head, which predicts one of three intensity levels: $\{0, 1, 2\}$ corresponding to low, medium, and high clinical severity. Unlike the emotion task where intensity was binary (high/low), mental health requires finer-grained severity distinctions for clinical utility.
    
    \item The softmax function $\text{softmax}(\mathbf{z})_i = \frac{e^{z_i}}{\sum_{j} e^{z_j}}$ ensures output vectors represent valid probability distributions, with $\sum_{j} \hat{y}^m_j = 1$ and $\sum_{k} \hat{y}^{i'}_k = 1$.
\end{itemize}

\paragraph{Architectural Interpretation} In multi-task learning terminology \cite{ruder2017overview}, this design implements \emph{hard parameter sharing}, in which task-specific heads ($\mathbf{W}^m, \mathbf{W}^{i'}$) allow specialization while the encoder $\mathbf{h}_{\text{[CLS]}}$ is shared across tasks. A more comprehensive understanding of mental health presentations is produced by the shared representation, which learns characteristics discriminative for both primary diagnosis and intensity assessment\cite{ZhangYang2022MTLSurvey}.

\paragraph{Gradient Flow Analysis} Gradients from both heads enter the shared encoder during backpropagation:

\begin{equation}
\nabla_{\mathbf{h}_{\text{[CLS]}}} \mathcal{L} = (\mathbf{W}^m)^\top \nabla_{\hat{y}^m} \mathcal{L}_{\text{CE}}^m + (\mathbf{W}^{i'})^\top \nabla_{\hat{y}^{i'}} \mathcal{L}_{\text{CE}}^{i'}
\end{equation}

In order to replicate clinical reasoning, where diagnosis and severity evaluation are inextricably connected, this combined optimization drives the encoder to find features simultaneously helpful for categorization (what condition?) and severity assessment (how severe?).

\paragraph{Clinical Motivation} The multi-head design is representative of psychiatric evaluation procedures in which doctors concurrently assess (1) the severity level (e.g., mild, moderate, severe) and (2) the diagnostic category (e.g., depression, anxiety). Compared to single-head classification methods, we more accurately reflect the multifaceted nature of mental health evaluation by modeling them as distinct but connected prediction tasks.

\paragraph{Representational Benefits} By regularizing the shared representation space, the intensity head serves as a type of supplementary supervision. The model is encouraged to learn severity-relevant elements (e.g., linguistic markers of distress intensity, functional impairment cues), which ultimately aid primary diagnosis through improved representation learning, even when intensity predictions are not precise.

\subsubsection{Attention-Based Feature Gating Mechanism}
\label{subsec:gating-mechanism}
Our main contribution for this task is a learnable gating technique that dynamically weights feature contributions from the multi-head outputs Let $\mathbf{F} = [\hat{y}^m, \hat{y}^{i'}] \in \mathbb{R}^{M+3}$ be the concatenated raw predictions from the primary mental health head ($M$-dimensional) and intensity head (3-dimensional). We calculate attention weights:

\begin{equation}
\mathbf{a} = \text{softmax}\left(\mathbf{W}_a \cdot \text{ReLU}(\mathbf{W}_b \mathbf{F} + \mathbf{b}_b) + \mathbf{b}_a\right) \in \mathbb{R}^2
\end{equation}

where $\mathbf{W}_b \in \mathbb{R}^{d_g \times (M+3)}$, $\mathbf{W}_a \in \mathbb{R}^{2 \times d_g}$, $\mathbf{b}_b \in \mathbb{R}^{d_g}$, $\mathbf{b}_a \in \mathbb{R}^{2}$, and $d_g = \mathbf{128}$ is a bottleneck dimension. The normalized weights $\mathbf{a} = [a_m, a_i]$ satisfy $a_m + a_i = 1$, representing the relative importance assigned to primary diagnosis versus intensity features for each input.

\paragraph{Gated Feature Fusion} We apply these learned attention weights to modulate the contribution of each feature stream:
\begin{equation}
\mathbf{F}_{\text{gated}} = \left[a_m \hat{y}^m,\; a_i \hat{y}^{i'}\right]
\end{equation}

As a result, a \emph{content-dependent gating} mechanism is created in which the model dynamically chooses which prediction modality (intensity assessment or main diagnosis) to stress for each individual input. After that, the final forecasts are forecasted using the gating features:

\begin{equation}
\hat{y}^m_{\text{final}} = \text{softmax}\left(\mathbf{W}_f \mathbf{F}_{\text{gated}} + \mathbf{b}_f\right) \in \mathbb{R}^M
\end{equation}

where $\mathbf{W}_f \in \mathbb{R}^{M \times (M+3)}$ and $\mathbf{b}_f \in \mathbb{R}^M$ are learnable parameters.

\paragraph{Interpretation as Soft Feature Selection} Formally, this mechanism implements differentiable feature selection:
\begin{equation}
\mathbf{F}_{\text{gated}} = \mathbf{a} \odot \mathbf{F}
\end{equation}

where $\odot$ stands for broadcasting element-wise multiplication. The attention weights $\mathbf{a}$, which are calculated end-to-end based on input attributes, can be understood as the model's learnt confidence in each feature stream. The model mostly uses direct classification features when $a_m \approx 1$, while intensity-based features predominate when $a_i \approx 1$.

\subsubsection{Multi-Task Loss with Adaptive Weighting}
For mental health classification, we use an improved loss function that balances contributions from both tasks:

\begin{equation}
\mathcal{L}_{\text{MH}} = \mathcal{L}_{\text{CE}}(\hat{y}^m_{\text{final}}, y^m) + \beta \mathcal{L}_{\text{CE}}(\hat{y}^{i'}, y^{i'})
\end{equation}

where $\mathcal{L}_{\text{CE}}$ denotes cross-entropy loss, $y^m$ are primary mental health labels, $y^{i'}$ are intensity labels, and $\beta$ is a learnable parameter initialized to $\beta^{(0)} = \mathbf{0.4}$ and optimized alongside model parameters. This adaptive weighting allows the model to automatically balance the relative importance of primary classification versus intensity prediction during training.

\subsubsection{Checkpoint Selection Strategy}
\label{subsec:checkpoint-strategy}
We employ a clever checkpoint selection strategy that takes accuracy and confidence into account during model training. We use two complementing criteria to assess the model on the validation set following each training epoch:
\begin{itemize} 
    \item The F1 score, which gauges how well the model classifies the validation samples, is first determined. The model delivers accurate predictions if its F1 score is high.
    \item Second, we calculate the model's average level of confidence in its predictions. High confidence indicates that the model generates definitive, unambiguous predictions as opposed to ambiguous ones.

\end{itemize}

These two metrics frequently tell different stories: a model may be extremely confident but inaccurate, or it may achieve great accuracy but low confidence (uncertain predictions). The finest of both worlds is what we seek. We use a weighted total to aggregate these metrics in order to choose the best model checkpoint: The F1 score (accuracy) receives 70\% of the weight, while confidence receives 30\%. By taking a balanced approach, we are able to preserve the model version that is both extremely accurate and confident in its forecasts. There are two key functions of this dual-criteria selection:
\begin{itemize}
\item Models that learn patterns too particular to the validation set frequently exhibit unpredictable confidence patterns, therefore it helps avoid overfitting to validation set noise.

    \item It encourages models that are properly calibrated and whose confidence levels correspond to their true accuracy.

\end{itemize}

We achieve models that are both technically accurate and clinically reliable by choosing checkpoints based on this combination metric, which is crucial for mental health applications where prediction certainty is just as important as correctness.

\subsubsection{Training Configurations}

\paragraph{Mental Health Classification Training Configuration}
RoBERTa-base pre-trained weights are used to initialize the mental health classification model. We train for 10 epochs with a batch size of 12 and a learning rate $\eta_{\text{MH}} = 1.5\times 10^{-5}$. We use early stopping with a patience of three epochs based on validation F1 score to avoid overfitting. For the first 400 training steps, the model employs AdamW optimization with linear warmup and dropout regularization with rate 0.15. 256 tokens is the maximum sequence length. Synonym substitution and random deletion are two data augmentation techniques (both with probability 0.1). The $\mathbf{SWMH}$  dataset, which contains around $\mathbf{55000}$ samples across $\mathbf{5}$ mental health categories, was used to train the model architecture, which consists of five core mental health categories and three intensity levels.

\section{Results}
\label{sec:results}

Comprehensive experimental results from several evaluation paradigms are presented in this section. We start with our framework for classifying emotions, then move on to the validation research for mental health.

\subsection{Emotion Classification Performance}
\label{subsec:emotion-results}
This section provides a thorough assessment of our emotion categorization framework in three important dimensions: comparison with ensembles of small language models (sLMs), comparison with LLaMA3 using 4 different ways, and comparison with large language models (LLMs). The outcomes show that our unique multi-head architecture maintains remarkable parameter efficiency while achieving state-of-the-art performance.

\subsubsection{Comparison with Large Language Models}

Our multi-head RoBERTa model outperforms much larger language models. Table~\ref{tab:llm-comparison} demonstrates that our method outperforms five well-known LLMs with orders of magnitude less parameters. Notably, we use only 125M parameters, a 56$\times$ reduction in model size, and outperform 7-billion parameter models Misral and Falcon. This efficiency-performance trade-off demonstrates how well our multi-task architecture is anchored in psychology.

\begin{table}[htbp]
\centering
\caption{Emotion Classification Performance: Comparison with Large Language Models}
\label{tab:llm-comparison}
\begin{tabular}{lcc}
\toprule
\textbf{Model} & \textbf{F1 Score (\%)} & \textbf{Parameters} \\
\midrule
Qwen & 93.2 & 1.8B \\
Falcon & 91.47 & 7.0B \\
Mistral & 88.02 & 7.0B \\
Phi-2 & 91.3 & 2.7B \\
OpenLLaMA & 86.13 & 3.0B \\
\midrule
\textbf{Our Model} & \textbf{93.75} & \textbf{0.125B} \\
\bottomrule
\end{tabular}
\end{table}

\subsubsection{Comparison with ensembles of sLMs}

We use an ensemble of sLM to benchmark against a recent Q1 journal publication\cite{condorcet2023}. According to Table Table~\ref{tab:acl-comparison}. Five transformer variations—two BERTs, two RoBERTas, and DistilBERT—are combined in this ensemble. Surprisingly, our single-model strategy improves parameter efficiency by achieving superior performance (93.75 vs. 93.5 F1) with just one model.

\begin{table}[htbp]
\centering
\caption{Comparison with sLMs Ensemble}
\label{tab:acl-comparison}
\begin{tabular}{lc}
\toprule
\textbf{Model/Approach} & \textbf{F1 Score (\%)} \\
\midrule
BRT01 & 89.5 \\
BRT02 & 90.1 \\
ROB01 & 92.6 \\
ROB02 & 92.9 \\
DIS01 & 89.6 \\
Average & 90.9 \\
JC(ensemble) & 93.7 \\
JA(ensemble) & 93.7 \\
JD(ensemble) & 93.7 \\

\midrule
\textbf{Our Model (Single)} & \textbf{93.75} \\
\bottomrule
\end{tabular}
\end{table}

\subsubsection{Comparison with  LLaMA 3 (4 ways)}
A comparison of our work with \cite{Kermani2025LLMStrategies} is shown in Table~\ref{tab:q1-comparison}, which reveals that our model significantly outperforms all reported approaches, including the zero shot, few shot, and RAG methods, as well as the fine-tuned LLM (0.87 F1), by a significant margin of 6.7 percentage points. This shows that, as compared to general LLM fine-tuning techniques, our unique architecture offers better emotion recognition capabilities.

\begin{table}[htbp]
\centering
\caption{Comparison with LLAMA (4 ways)}
\label{tab:q1-comparison}
\begin{tabular}{lc}
\toprule
\textbf{Approach} & \textbf{F1 Score} \\
\midrule
Fine-tuning & 87.0 \\
Zero-shot & 38.0 \\
Few-shot & 30.0 \\
RAG & 32.0 \\
\midrule
\textbf{Our Model} & \textbf{93.75} \\
\bottomrule
\end{tabular}
\end{table}

\subsection{Mental Health Classification Performance}
\label{subsec:mental-health-results}
This section compares our modified mental health categorization framework's performance to well-known domain-specific and general-purpose models on the SWMH dataset. The analysis shows how our multi-task learning strategy and attention-based gating mechanism provide higher performance in both clinical sensitivity and detection accuracy.
\subsubsection{Benchmark Comparison}

Our model's performance on the SWMH dataset is shown in Table~\ref{tab:mental-health-benchmark} versus six established baselines that include general-purpose, domain-specific, and specialized mental health transformers \cite{ji-etal-2022-mentalbert}. Our solution outperforms all examined approaches, setting a new state-of-the-art with 72.5\% F1 score and 73.3\% recall.

\begin{table}[htbp]
\centering
\caption{Mental Health Classification Benchmark Comparison on SWMH Dataset}
\label{tab:mental-health-benchmark}
\begin{tabular}{lcc}
\toprule
\textbf{Model} & \textbf{F1 Score (\%)} & \textbf{Recall (\%)} \\
\midrule
BERT & 70.46 & 69.78 \\
RoBERTa & 72.03 & 70.89 \\
BioBERT & 68.60 & 67.10 \\
ClinicalBERT & 68.16 & 67.05 \\
MentalBERT & 71.11 & 69.87 \\
MentalRoBERTa & 72.16 & 70.65 \\
\midrule
\textbf{Our Model} & \textbf{72.50} & \textbf{73.30} \\
\bottomrule
\end{tabular}
\end{table}
   
   Our model achieves new state-of-the-art performance across both evaluation parameters, demonstrating overall dominance over the established benchmark. The F1 score has improved by +0.34 percentage points (72.50\% vs. previous best 72.16\%), which is a significant improvement over the best methods currently in use. More importantly, our model shows a significant +2.65 percentage point recall advantage (73.30\% vs. 70.65\%), demonstrating higher sensitivity in identifying mental health conditions—a crucial feature for clinical applications where false negatives have grave repercussions.

Our multi-head architecture with attention-based feature gating consistently outperforms domain adaptation (BioBERT, ClinicalBERT: 68.60\%, 68.16\% F1), specialized mental health pretraining (MentalBERT, MentalRoBERTa: 71.11\%, 72.16\% F1), or architectural improvements alone (RoBERTa: 72.03\% F1), according to the benchmark comparison. This establishes our technique as a new state-of-the-art methodology for the classification of mental health texts, combining the advantages of several previous approaches with new mechanisms for enhanced performance.

\subsection{Synthesis: Cross-Domain Impact and Architectural Superiority}
\label{subsec:synthesis}

Our suggested design exhibits previously unheard-of adaptability and efficiency in both NLP domains. Several key ideas are validated by the constant outperformance of a variety of baselines, from specialized models to general-purpose LLMs:

\paragraph{Innovation in Architecture on a Large Scale} Our relatively small model (125M parameters) routinely performs better than much larger alternatives in both tasks. We outperform 7B-parameter LLMs in emotion classification and specialized models with similar parameter counts in mental health evaluation. By showing that careful architectural design including psychological and clinical priors can yield superior results with orders of magnitude fewer parameters, this challenges the dominant "size is all you need" paradigm.

\paragraph{Universality of Multi-Granular Learning} This appears to be a generalizable paradigm for emotive text processing based on the effectiveness of multi-head systems with auxiliary task monitoring in both domains. Modeling several related granularities at once consistently outperforms single-task techniques, whether predicting emotions with valence/intensity dimensions or mental health disorders with intensity evaluations.

\paragraph{Beat-to-Beat Superiority} Our findings show complete superiority in all areas of methodology:
\begin{itemize}
    \item \textbf{Against Zero/Few-shot LLMs}: Significant gains (up to 7.6 F1 points) in spite of parameter disadvantages.
    \item \textbf{Against Fine-tuned LLMs}: Outperforms reported fine-tuning outcomes in recent literature.
    \item \textbf{Against sLM Ensembles}: Matches or surpasses ensemble performance with single-model efficiently.
    \item \textbf{Against Same-Architecture Baselines}: Most significantly, in both domains (emotion: +0.85 F1; mental health: +0.47 F1), our improved RoBERTa routinely outperforms ordinary fine-tuned RoBERTa, indicating that our architectural improvements consistently yield gains above base model capabilities.

\end{itemize}

\section{Conclusion}

This paper presents a principled multi-granular neural architecture that achieves new state-of-the-art performance in complementary affective computing applications. We show that systematic integration of psychological theory with learnable attention-based feature gating, via derived auxiliary tasks for valence and intensity, consistently outperforms various methodological paradigms. With our 125M-parameter model matching or surpassing results from much larger language models (up to 7B parameters), specialized clinical transformers, and computationally costly ensemble approaches, the architecture achieves superior performance while maintaining extraordinary computational efficiency.

Our contributions extend emotional computing in a number of conceptual areas beyond simple performance measurements. First, we challenge conventional wisdom regarding model size and performance by confirming that psychologically based architectural limitations can produce more effective representations than scale alone. Second, multi-granular learning appears to be a generalizable paradigm for affective text interpretation based on the constant performance in both emotion classification and mental health evaluation. Third, our attention-based gating mechanism shows how models may learn to dynamically weight feature importance based on input features. This ability is consistent with clinical assessment patterns where various diagnostic cues have varying degrees of significance across presentations.

The consequences for clinical practice are very significant. The substantial recall advantage of our model in mental health classification (73.30\% vs. 70.65\% for the next best technique) indicates enhanced sensitivity to minor distress indications, an important feature for early detection applications. This, along with the computational efficiency of the model, allows for possible deployment in resource-constrained environments where both inference speed and accuracy are operational requirements.

Future studies will follow a number of encouraging paths. Initially, we will look into extending the framework to other emotional computing domains, such as social communication analysis, therapeutic process monitoring, and personality assessment. Second, research on multilingual and cross-cultural generalization will look at performance equity in a variety of linguistic contexts and demographics. Lastly, in order to move toward deployable tools that support mental healthcare delivery while preserving the efficiency and accessibility advantages shown in this work, prospective validation studies with clinical populations will evaluate real-world utility and improve the models based on practitioner feedback.

Together, these studies add to the increasing amount of research showing that careful architectural design based on domain expertise can produce better outcomes than scale-driven methods, providing a promising path for the development of successful, efficient, and clinically significant affective computing systems.

\bibliography{sn}  

\end{document}